\documentclass[journal]{IEEEtran}

\usepackage{xcolor,soul,framed,caption}

\colorlet{shadecolor}{yellow}
\usepackage[pdftex]{graphicx}
\graphicspath{{../pdf/}{../jpeg/}}
\DeclareGraphicsExtensions{.pdf,.jpeg,.png}

\usepackage{threeparttable}
\usepackage{algorithm}
\usepackage{algpseudocode}
\usepackage{booktabs}
\usepackage{siunitx}
\usepackage{multirow}
\usepackage[cmex10]{amsmath}
\usepackage{amssymb}
\usepackage{bbm}
\usepackage{array}
\usepackage{mdwmath}
\usepackage{mdwtab}
\usepackage{eqparbox}
\usepackage{url}
\usepackage{hyperref}
\hypersetup{
    colorlinks=true,
    linkcolor=blue,
    filecolor=magenta,      
    urlcolor=cyan
}
\usepackage[capitalize]{cleveref}
\crefname{section}{Sec.}{Secs.}
\Crefname{section}{Section}{Sections}
\Crefname{table}{Table}{Tables}
\crefname{table}{Tab.}{Tabs.}
    
\begin{document}
\bstctlcite{IEEEexample:BSTcontrol}
\title{A Hierarchical Hybrid Learning Framework for Multi-agent Trajectory Prediction}

\author{Yujun Jiao$^{1 \dagger}$ ~Mingze Miao$^{1 \dagger}$ ~Zhishuai Yin$^{1 *}$ ~~Chunyuan Lei$^1$ ~Xu Zhu$^1$ ~Linzhen Nie$^1$ ~Bo Tao$^1$
	\thanks{$\dagger$ These authors contributed equally to this work.}
	\thanks{* Corresponding author.}\\
	$^1$Wuhan University of Technology\\
	\{yjiao, mingze.miao, zyin, chunyuan.lei, xu.zhu, linzhen\_nie, taobo2020\}@whut.edu.cn\\
}
\maketitle
\begin{abstract}
Accurate and robust trajectory prediction of neighboring agents is critical for autonomous vehicles traversing in complex scenes. Most methods proposed in recent years are deep learning-based due to their strength in encoding complex interactions. However, unplausible predictions are often generated since they rely heavily on past observations and cannot effectively capture the transient and contingency interactions from sparse samples. In this paper, we propose a hierarchical hybrid framework of deep learning (DL) and reinforcement learning (RL) for multi-agent trajectory prediction, to cope with the challenge of predicting motions shaped by multi-scale interactions. In the DL stage, the traffic scene is divided into multiple intermediate-scale heterogenous graphs based on which Transformer-style GNNs are adopted to encode heterogenous interactions at intermediate and global levels. In the RL stage, we divide the traffic scene into local sub-scenes utilizing the key future points predicted in the DL stage. To emulate the motion planning procedure so as to produce trajectory predictions, a Transformer-based Proximal Policy Optimization (PPO) incorporated with a vehicle kinematics model is devised to plan motions under the dominant influence of microscopic interactions. A multi-objective reward is designed to balance between agent-centric accuracy and scene-wise compatibility. Experimental results show that our proposal matches the state-of-the-arts on the Argoverse forecasting benchmark. It's also revealed by the visualized results that the hierarchical learning framework captures the multi-scale interactions and improves the feasibility and compliance of the predicted trajectories.
\end{abstract}

\IEEEpeerreviewmaketitle

\section{Introduction}

\IEEEPARstart{P}{redicting} the future trajectories of surrounding traffic participants, or agents, is crucial to contingency motion planning of autonomous vehicles in complex driving scenes. However, the intertwining of heterogenous behavioral strategies and dynamic driving circumstances, which both are influenced by numerous stochastic factors and constrained by various sets of rules, poses tremendous challenges to identifying the complex interactive relations between traffic participants and the scene, therefore causes great difficulties in predicting the motion of each participant individually and even more so jointly. 

Traditional approaches employ physics-based or maneuver-based models, devised based on crafted rules or logics, to predict trajectories \cite{Xie2018,althoff2009model,sorstedt2011new} . The drawback is that they’re incapable of accurately modeling the heterogenous behavioral strategies of agents in arbitrary scenarios. Modern deep learning (DL)-based approaches explore from a different perspective in the sense that the prediction of trajectories is accomplished through learning motion trends from abundant observational data of behavioral phenomena. Most of these works are structured in the pattern of ``Encoder-Decoder". First, the encoder processes the features of dynamic and static traffic elements, e.g. agents and road lanelets, as sequences \cite{Yuan2021,Ngiam2021}, graphs \cite{Liang2020,Li2021}, images \cite{phan2020covernet,Park2020} and point clouds \cite{Ye2021}. Subsequently, different approaches are proposed to model the interactions between agents the scene, including pooling mechanisms \cite{Alahi2016,Gupta2018,Deo_CS}, graph neural networks (GNN) \cite{Gao2020,Gu2021,Zheng2021,Li2021,TNT}, attention mechanisms \cite{Gilles2022,Deo_CS,Nayakanti2022,Song2021,Kim2021}, and dual representation space learning \cite{Ye2021}. Finally, generative models such as multi-layer perceptron (MLP) based Gaussian Mixed Model(GMM) \cite{chai2019multipath}, conditional variational auto-encoders (CVAEs) \cite{lee2017desire,Yuan2021} and generative adversarial networks (GANs) \cite{Choi2022,Gupta2018,Zhao2019}, are used to decode the obtained features and produce multi-modal motion predictions to take the behavioral uncertainties into account. Recent efforts have also been devoted to mitigate conflicts of trajectory predictions for multiple agents, through joint predictions \cite{Gilles2021,Ngiam2021,Sun2022,chen2022scept}.
\begin{figure*}[t]
	\centering
	\includegraphics[width=0.95\linewidth]{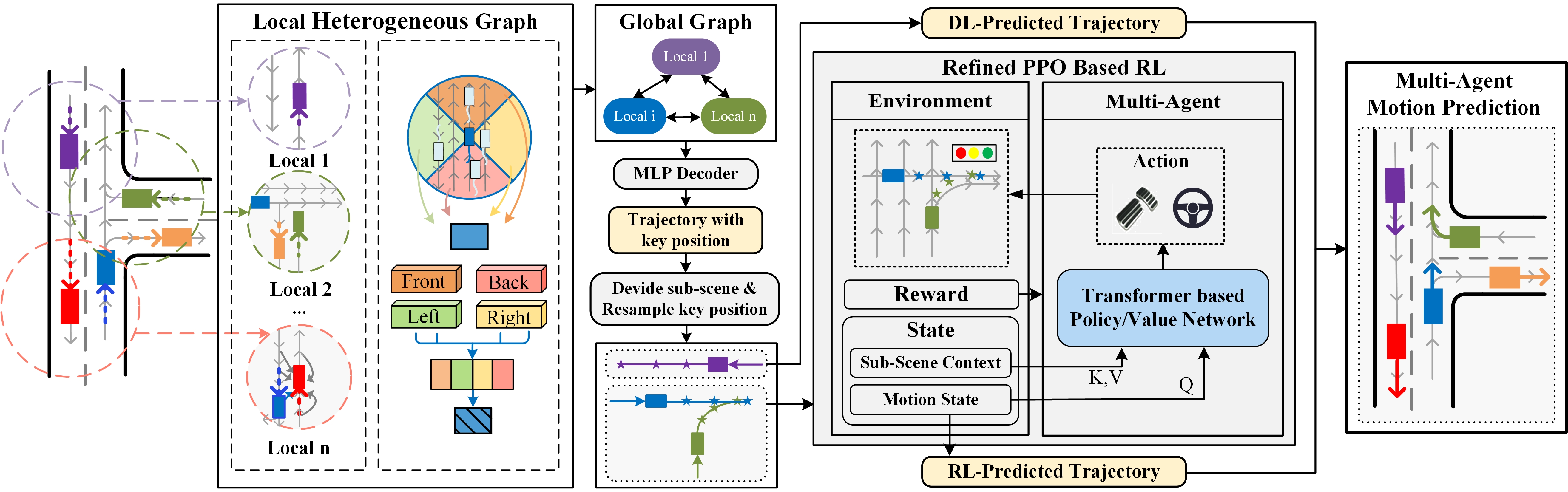}
	\caption{Overview of the hybrid framework.}
	\label{fig:overall}
\end{figure*}

Predicting trajectories, we argue, is in fact the process of interpreting and modeling the probabilistic motion planning procedure of the target agent, driven simultaneously by the agent's ambiguous multi-modal intentions and multi-scale interactions with the scene. Deep learning-based methods are effective in modeling interactions at the intermediate spatial-temporal scale and reasoning about multi-modal intentions. However, it's neither theoretically grounded nor effective to model the microscopic contingency motion planning procedure as a sampling task relying primarily on past observations. As a result, the DL methods could easily fail to generate feasible predictions conforming to traffic rules and contextual constraints in highly interactive scenes. 

Moreover, unlike low-speed robotics, vehicles traveling at high speed would almost never fully execute the trajectory planned in dense traffic. In fact, only about a few hundred milliseconds of the planned motion will be executed by the agent before its trajectory is replanned with updated knowledge about the surrounding scene \cite{da2022comprehensive}. Two conclusions could be drawn from this fact: First, from the perspective of the target agent, its motion plans although are partially determined by its historical states as well as the intermediate-scale scene context, they are often also stimulated responses to its instantaneous interactions with the microscopic-level local scene, therefore are outcomes of an optimal decision-making process with complex and subtle factors taken into account. Modeling such a process into a deep learning task based primarily on encoded historical states and intermediate-scale scene contexts would be intrinsically deficient in capturing transient yet critical interactions. Second, from the perspective of the ego-agent, which is the very consumer of the prediction results, focusing on enhancing the long-term or average prediction accuracy isn't necessarily the only valid goal as the ego-agent is also concerned with the short-term uncertainty of surrounding agents’ behaviors, for the purpose of safe contingency ego-motion planning.

In this study, we propose a hierarchical hybrid framework of deep learning and reinforcement learning (RL) for multi-agent trajectory prediction, to cope with the challenge of predicting motions driven concurrently by multi-modal intentions and multi-scale interactions. In the first DL-based stage, we decompose the overall scene into multiple intermediate-scale heterogenous graphs and introduce Transformer-style GNNs to encode interactions among agents and road lanelets at both intermediate and global levels. The outputs of the DL stage are predicted key future positions of agents. A simple yet efficient calibration process is implemented to check the compliance of the predicted key positions to contextual constraints and perform rectifications if not. In the second RL-based stage, we divide the overall scene into microscale local sub-scenes based on the predicted key positions. A Proximal Policy Optimization (PPO)-based RL model is devised to explore optimal behavioral strategies under the influence of instantaneous interactive relations within the sub-scene. Based on the multi-objective reward that requires all agents to traverse through their key positions collision-free at each time step, a balance between agent-centric accuracy and scene-wise compatibility is naturally achieved.

The main contributions of this study are outlined as follows:
\begin{enumerate}
\item We propose a novel hierarchical hybrid learning framework to decompose trajectory prediction into DL-based key position prediction and RL-based motion plan, exploiting the strength of each stage to capture the multi-scale interaction that collectively shape the motions of the agents.
\item A heterogeneous graph Transformer is constructed to encode the heterogenous relational features between agents in the intermediate-scale scene to predict the key future positions, which are subsequently used to define microscopic sub-scenes as focalized local environment for the RL process.  
\item We employ the RL framework to predict trajectories between key positions in the form of motion planning. As a result, trajectory prediction is more robust to stochastic circumstances since microscopic and transient interactions are captured. Scene-level compatibility also comes naturally to prediction results with the balanced multi-objective reward.
\item The proposed framework is evaluated on the Argoverse Forecasting Dataset, showing state-of-the-art (SOTA) performance. The visualized results prove the strength of our proposal in producing feasible and plausible trajectory predictions in highly interactive scenes.
\end{enumerate}
\section{Related Work}
\label{sec2}
\subsection{Scene Context and Interaction Encoding}
Modeling interactions among various types of traffic elements with awareness of the context, lays the foundation for trajectory prediction. Prompted by successes of Convolutional Neural Networks (CNNs) in other vision-based tasks, pioneering DL-based methods \cite{phan2020covernet,HOME,djuric2020uncertainty} employ CNNs to learn scene context and spatial-temporal interactive features from rasterized driving scenes, centered around the target agent from the Bird’s Eye View (BEV). Attention mechanisms are often involved in adaptively fusing features \cite{Deo_CS,Kim2021}. More recent work adopts Graph Neural Networks (GNNs) to encode historical states of agents and road elements on the basis of the constructed scene graphs whose nodes are traffic elements interconnected with edges representing interactions among them \cite{Gao2020,Liang2020,Zeng2021,Zhou2022,jia2022hdgt}. HiVT \cite{Zhou2022} encodes spatial-temporal information of the scene by decomposing the task hierarchically into local context encoding and global interaction modeling, both with Transformers as the core. HDGT \cite{jia2022hdgt} constructs the scene as heterogenous graphs composed of different types of nodes and edges. A few Transformers are then stacked to deal with the heterogeneity of the inputs. UNIN \cite{Zheng2021} proposes hierarchical graph attention modules to obtain interactions at the category level and agent level.
\subsection{Motion Prediction}
Although it’s acknowledged as an essential issue to ensure compatibility of predicted trajectories of multiple agents, most methods are in fact agent-centric. To produce multi-modal predictions, early works employ generative models such as CVAEs \cite{lee2017desire,Yuan2021} and GANs \cite{Choi2022,Gupta2018,Zhao2019} to predict future trajectories via stochastic sampling from the latent distribution of the conditional inputs. A more recently adopted approach is to predict the multi-modal goals and regress the trajectories conditioned on the goals \cite{Gilles2022,Gu2021,TNT,Zeng2021,HOME}. To avoid having to devise hand-crafted algorithms, KEMP \cite{Lu2022} proposes to generalize the goal-based work by introducing an intermediate task to predict key frames that trace out the general direction of the trajectory.

A few exploratory works have directed the focus onto joint prediction of multiple agents’ trajectories \cite{Gilles2021,Ngiam2021,Sun2022,chen2022scept}. THOMAS \cite{Gilles2021} proposes a scene-adaptive method to recombine marginal modalities of each agent to produce scene-level consistent predictions, based on a matching score that determines the weight of each modality in the overall scene. The matching score is calculated via cross-attention and therefore the computational complexity increases quadratically with the number of agents. M2I \cite{Sun2022} classifies the agents into influencers and reactors, whose trajectories are predicted with a marginal prediction model and conditional prediction model, respectively. The conditional prediction is conditioned on the predicted trajectories of the influence, so that overlaps of trajectories between multiple agents are alleviated.
\subsection{Deep Reinforcement Learning-based Motion Planning}
Deep Reinforcement Learning (DRL) has been extensively applied in modeling the procedure of motion planning of intelligent agents and has shown great potentials \cite{Saxena2020,Strudel2020,li2019sarl,ye2020automated}. In the framework of DRL, environmental features extracted by the DL-based module are passed to the RL procedure to explore optimal strategies iteratively. Existing DRL methods fall into the categories of value-based methods, such as Deep Q-Network (DQN) \cite{Mnih2015} and its variants \cite{Hessel2018}, and policy-based methods, such as PPO \cite{schulman2017proximal}, Soft Actor-Critic (SAC) \cite{Haarnoja2018}, etc.

The strength of DRL lies in the ability to learn dynamic and implicit interactive relations among agents. Our study models trajectory prediction within the local sub-scene as a DRL-based motion planning process. To the best of knowledge, it’s the first attempt of predicting trajectories by resembling the motion planning process in the framework of DRL.
\section{Approach}
\subsection{Overall}
Our method encodes the interactions at three levels: global, intermediate and local with a hierarchical hybrid framework consisting of a DL-based and a RL-based stage, as shown in Fig. \ref{fig:overall}. The DL-based stage encodes interactions at both global and intermediate levels, producing key future positions as outputs. The RL-based stage constructs local sub-scenes based on the predicted key positions and models the probabilistic motion planning procedure of the target agent influenced by microscopic interactions within each divided sub-scene. It also serves the role as the implicit decoder by producing multi-modal trajectories of multiple agents as results of motion planning.
\subsection{Heterogeneous Graph Transformer}
\label{DL}
\begin{figure}
	\centering
	\includegraphics[width=0.8\linewidth]{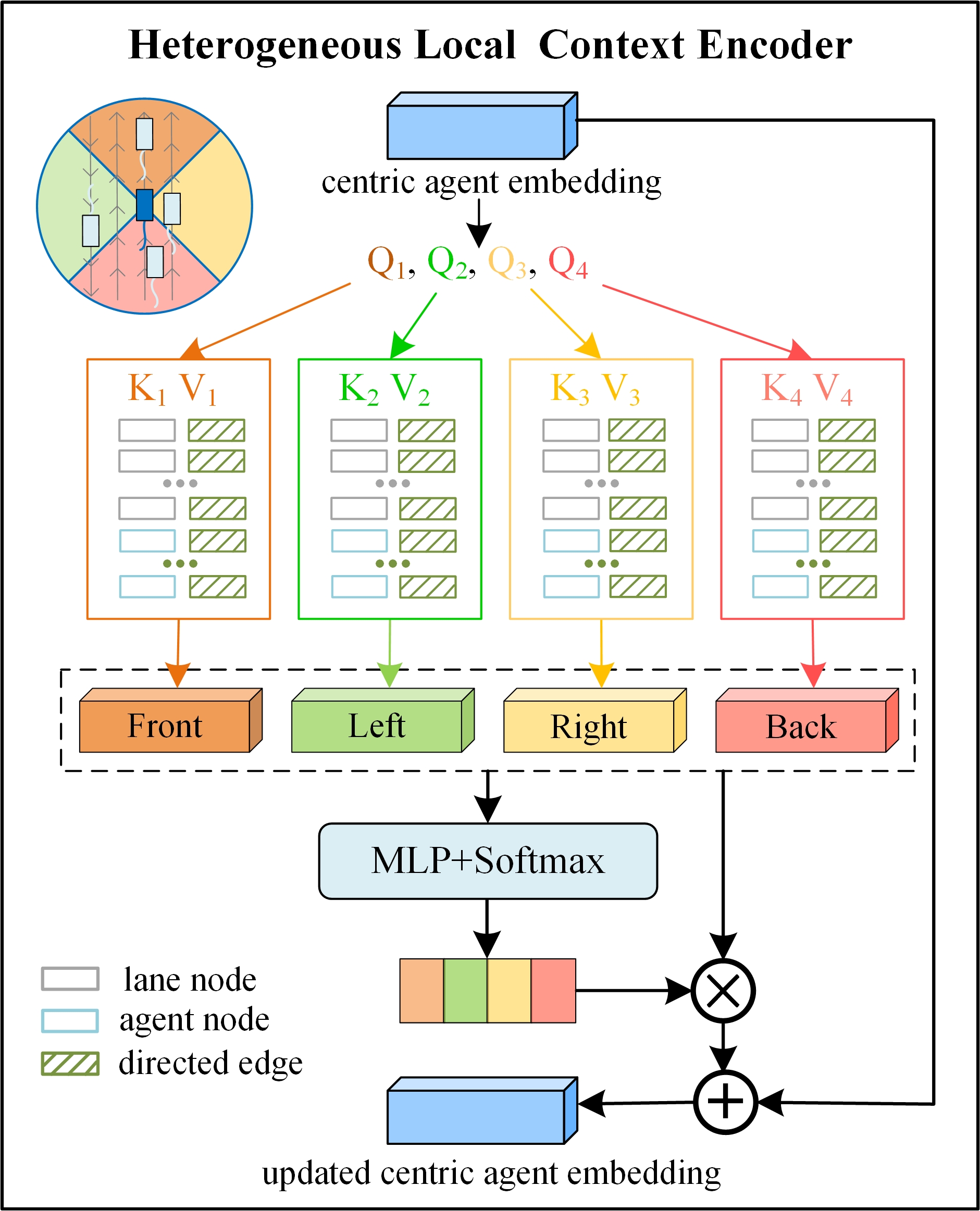}
	\caption{Heterogenous local context encoder.}
	\label{fig:Fig2_DL}
\end{figure}
To expedite scene representation and interaction encoding, without losing the ability to capture the transmission of message at the global level, we adopt the hierarchical framework similar to HiVT \cite{Zhou2022} to encode the scene at both intermediate and global levels, with special focus on the former part. Instead of encoding heterogenous interactions sequentially, we construct agent-centric heterogenous sub-graphs to represent the surrounding scene within an empirical distance to a selected target agent. The motivation is that heterogenous traffic elements are in fact interacting with the central agent simultaneously in a tight coupling manner. Besides different types of traffic elements, we introduce another level of heterogeneity by discriminating interactions coming from the same type of elements but in different directions relative to the heading of the target agent. To that end, a local heterogeneous graph with four types of edges and two types of nodes was constructed based on four directions (front, left, right, and back) and two traffic elements (lane and agent).

The overall directed heterogeneous graph is denoted as $G=\left\{ G_s \mid \forall s \in \left\{1, \dots ,S\right\}\right\}$, where $G_s$ is the heterogeneous sub-graph of the local scene $s$. For the local scene $s$ comprises $M$ types of $N$ heterogenous traffic elements, the sub-graph is defined as $G_s=\left\{V_s,E_s\right\}$, where $V_s=\left\{v_n \mid \forall n \in \left\{1, \dots ,N \right\} \right\}$ is the set of $N$ nodes, and $E_s=\left\{e_{ij} \mid \forall i,j \in \left\{1, \dots ,N\right\}\right\}$ is the set of directed edges. The attribute of nodes is defined as $v_n=\left\{f_v^n\right.$, OneHotEnconding $\left.(T)\right\}$, where $f_v^n$ and $T$ denotes the embedding feature and type of the node. The attribute of edges is defined as $e_{ij}={f_e^{ij}}$, where $f_e^{ij}$ denotes the embedding feature.

The vectorized road lanelets in HD-Map and the historical states of agents are respectively fed into the linear embedding layer and a Temporal Transformer to obtain the embedding features $f_v^n$ of heterogenous nodes. For each edge, we obtain the initial embedding features $f_e^{ij}$ from relative position between nodes with an MLP.

Based on the constructed hetero-graph, a Transformer-style GNN is adopted to encode the intermediate-level scene context, which is decoupled in each layer into two phases: aggregating features of adjacent source nodes and combining the aggregated features with the central node’s features from the previous layer. Specifically, in the phase of aggregation, for each quadrant, we stack a Transformer to aggregate features of adjacent heterogenous source nodes, by setting the vectorized feature of the central agent node $v_{centre}$ as $Q_d$, and that of the source node $v_{source}^d$ and the edge $v_{centre}$ leaving $v_{source}^d$ as $K_d$ and $V_d$, as expressed in Eq.(\ref{eq:aggr}). 
\begin{equation}
	\begin{gathered}
		Q_d=v_{\text {centre}} W_d^Q, \\
		K_d=\operatorname{concat}\left[v_{\text {source}}^d, e\right] W_d^K, \\
		V_d=\operatorname{concat}\left[v_{\text {source}}^d, e\right] W_d^V, \\
		v_{\text {aggr}}^d=\operatorname{softmax}\left(\frac{Q_d K_d^T}{\sqrt{d_k}}\right) V_d,
	\end{gathered}
	\label{eq:aggr}
\end{equation}
\noindent
where $d_k$ is the dimension of the query and key embeddings, and $W_d^Q$, $W_d^K$, $W_d^V$ are learnable matrices.

Subsequently, in the phase of combination, we update the node feature of the central agent by combining aggregated features $v_{aggr}=\left\{v_{aggr}^d \mid \forall d \in\{1, \cdots, 4\}\right\}$ in different directions, through weighted element-wise addition, as expressed in Eq.(\ref{eq:addition}). The weight is adaptively learned with an attention module, to take into account the fact that interactions coming from different directions affect the behavioral strategies of the central agent differently.
\begin{equation}
	\begin{aligned}
		& \omega=\operatorname{softmax}\left(M L P_{d_m \rightarrow 1}\left(v_{\text {aggr }}\right)\right) \\
		& \hat{v}_{\text {centre }}=\sum_{d=1}^4 \omega_d * v_{\text {aggr }}^d+v_{\text {centre }}
	\end{aligned}
	\label{eq:addition}
\end{equation}
Where $d_m$ is the dimension of the hidden layer, $w =\left\{w_d \mid \forall d \in\{1, \cdots, 4\}\right\}$.

Furthermore, we introduce a global graph Transformer to allow message passing across local graphs. Finally, an MLP decoder is attached to predict multi-modal key future positions for all agents.
\subsection{The Refined PPO based RL Model}
\label{RL}
\begin{figure}
	\centering
	\includegraphics[width=0.9\linewidth]{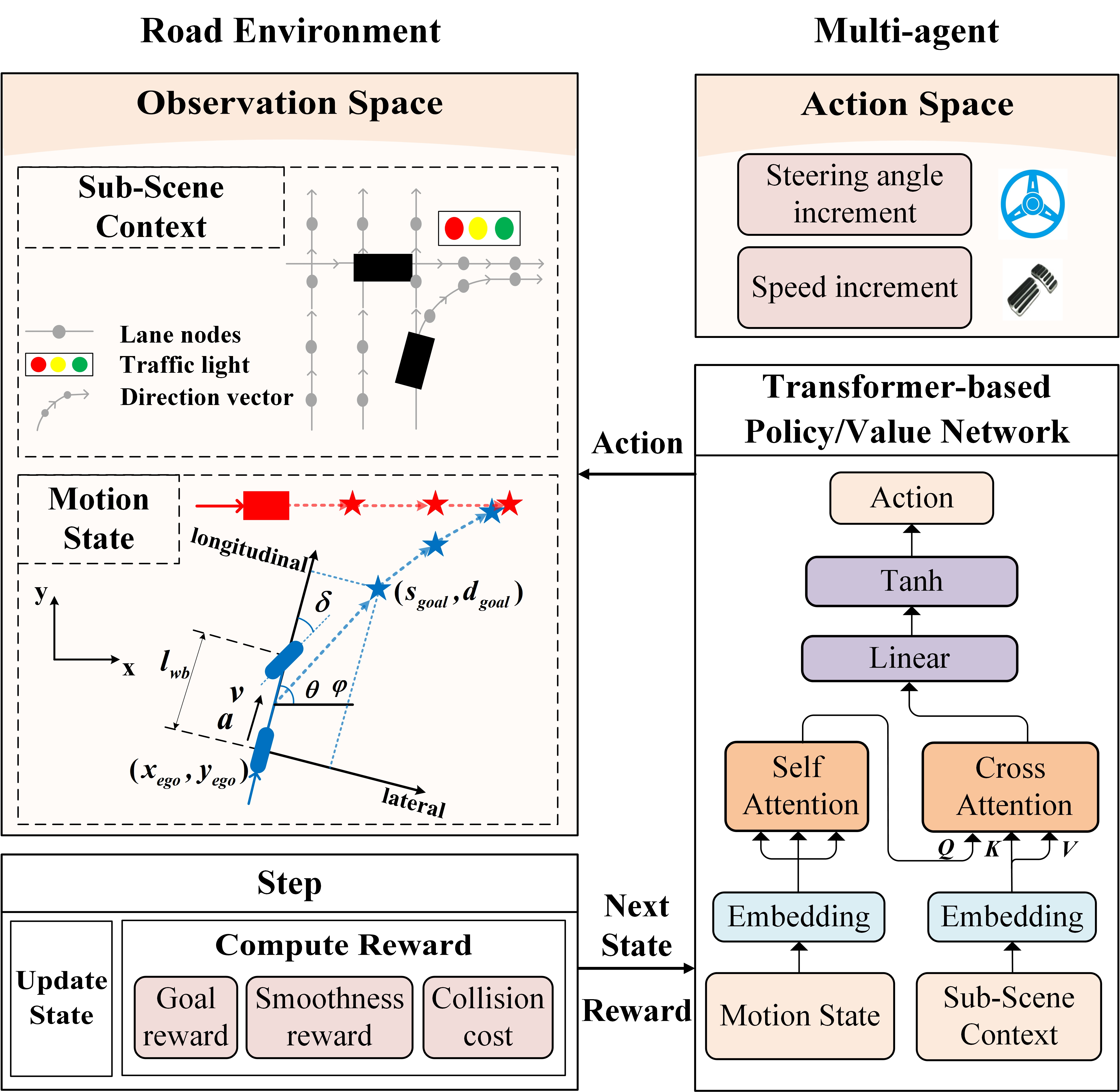}
	\caption{The refined interactive PPO-based RL model.}
	\label{fig:Fig3_RL}
\end{figure}
In the RL stage, to focus the RL model on the local environment in which the agent learns a collision-free strategy, we divide the global scene into a series of local sub-scenes based on the key positions predicted for each agent in the previous stage. The overall diagram of the RL stage is presented in Fig. \ref{fig:Fig3_RL}.
\begin{table*}
	\begin{center}
		\resizebox{1.2\columnwidth}{!}{
			\begin{tabular}{|l|c|c|c|c|}
				\hline
				Method                                              & minADE          & minFDE          & MR                      & DAC           \\
				\hline\hline
				LaneGCN\cite{Liang2020}                             & 0.9038          & 1.4526          & \textbf{0.1232}         & \textbf{0.9903}  \\
				mmTransformer\cite{liu2021multimodal}               & 0.8346          & 1.3383          & 0.1540                  & 0.9842           \\
				TNT\cite{TNT}                                       & 0.9097          & 1.4457          & 0.1255                  & 0.9889           \\
				DenseTNT\cite{Gu2021}                           	& 0.8817          & 1.2858          & 0.1285                  & 0.9875           \\
				TPCN\cite{Ye2021}                                   & 0.8153          & 1.2442          & 0.1333                  & 0.9884           \\
				HiVT-128\cite{Zhou2022}                             & \textbf{0.7735} & \textbf{1.1693} & 0.1267                  & 0.9888         \\
				\hline\hline
				Ours                                                & \textbf{0.8046} & \textbf{1.2461} & \textbf{0.1392}         & \textbf{0.9905}   \\
				\hline
		\end{tabular}}
	\end{center}
	\caption{\centering{Comparison with published state-of-the-art methods on Argoverse Forcasting public leaderboard.}}
	\label{tab_1}
\end{table*}

\textbf{Vehicle Model and Action Space}
 
Benefited from transforming trajectory prediction into the process of motion planning, we incorporate the single-track vehicle kinematic model into the RL stage, so that the planned motions, or the predicted trajectories in the context of this study, are compliant to vehicle kinematics. Since vehicle motion control comprises longitudinal and lateral dimensions, we define a 2-dimensional action space consisting of a steering angle increment and a speed increment, both of which are conforming to Gaussian distribution.

\textbf{Observation Space}

For convenience, we denote the lanelets that are associated with predicted key positions as key lanelets. To involve all traffic elements related with the target agent, we define the local sub-scene based on both prior and posterior knowledge about the driving scene, to include: 1) topologically related elements, namely lanelets that are directly connected to the key lanelet in the HD-map, along with agents on them; 2) interactive elements, namely agents whose predicted key positions are within an empirical distance from those of the target agent, along with lanelets beneath them. 

Both the sub-scene context and the agent's motion states are formulated into the observation space $O_i$. To produce the sub-scene context, each lane node within the lanelet, denoted as $N_p$ to avoid confusion with the node of the aforementioned GNN, are augmented to include the road boundary $b$, the traffic direction attribute $d_t$, and the passable state controlled by the traffic light $p$. Subsequently, a 1x1 convolutional layer is adopted to map the sequence of discrete lanes nodes to the sub-scene context, denoted as \{$N_p$\}. As for the motion state of the ego vehicle, we concatenate the speed $v$, the longitudinal acceleration $a_{long}$ (the lateral acceleration is trivial and neglected), the steering angle $\delta$, the yaw rate $\varphi$, the position $(s_{goal},d_{goal})$ of the key position goal in the vehicle coordinate system, the remaining time to reach the goal $t$ , the heading angle $\theta$ as well as the position $(x_{ego},y_{ego})$ of the ego vehicle in the global coordinate system (for multi-vehicle interaction) , and fed it into a 1x1 convolutional layer $\mathcal{F}$ to produce a motion state embedding vector $M_i$ in the same dimension as the sub-scene context embedding. The constructed observation space $O_i$ is expressed as in Eq.(\ref{eq:4}).
\begin{equation}
	\begin{footnotesize}
		O_i=\left\{\begin{array}{c}
			\left\{N_p\right\}=\left\{\mathcal{F}\left(\operatorname{concat}\left(x, y, d_t, o, p\right)\right)\right\} \\
			M_i=\mathcal{F}\left(\operatorname{concat}\left(v, a_{long}, \varphi, s_{goal}, d_{goal}, \theta, x_{ego}, y_{ego}\right)\right.
		\end{array}\right.
		\label{eq:4}
	\end{footnotesize}
\end{equation}

\textbf{The Refined Interactive PPO}

In our study, a Transformer-based module is introduced to replace the regular neural network (NN) layers in the vanilla-PPO algorithm, as shown in Fig. \ref{fig:Fig3_RL}. We set the agent’s motion state embedding as $Q$, the sub-scene context embedding as $K$ and $V$. The substitution of regular NN with Transformer isn’t a naive attempt for two reasons. First, Transformer natively supports the robust and efficient implementation of cross-attention which in our case is adopted to learn the rapidly evolving interactive relations between the agent and the sub-scene context. Second, Transformer natively accommodates to arbitrary sub-scenes in the sense that it is adaptable to an observation space with arbitrary dimensions, determined by the number of traffic elements involved in the sub-scene.

\textbf{Scene Compatibility-aware Reward Function}

To resolve the conflicts between agent-centric methods which focus overwhelmingly on prediction accuracy of each individual agent, and scene-wise methods driven dominantly by the goal of eliminating incompatible predictions among multiple agents, we design a multi-objective reward function to balance between the goal of maximizing the reward for each individual agent and maximizing the global mean reward of all agents presented in the sub-scene. The reward function of each agent is designed, as expressed in Eq.(\ref{eq:reward}), to include the rewards from both individual and collective perspectives, with a slight bias toward the former one.

First of all, every time the agent executes an action, it gets a goal reward $R_{goal}$, which is obtained by computing the heat value of the current position in a heat map generated based on the Standard Normal distribution centered at the target position, so as to solve the problem of sparse rewards before the agent reaches the destination. At the step that should reach the key position, this reward is multiplied by a coefficient greater than 1 to drive the agent toward the key position. Second, in order to emulate the objective of driving comfort when human drivers plan the motion, we set a smoothness reward $R_{smooth}$, which restricts the longitudinal acceleration and front wheel angle increment in the action space, respectively. Finally, we set the collision penalty $R_{collision}$, to avoid potential collisions with other agents or road elements.
\begin{equation}
	\begin{small}
		\begin{gathered}
			R_{\text {goal }}(t)=\mathcal{N}\left(p_t, \mu=p_{goal}, \sigma\right) \\
			R_{\text {smooth }}(t)=\left\{\begin{array}{ll}
				-\mathcal{N}\left(a_{long}, \mu=0, \sigma\right) \\
				-\mathcal{N}(\delta, \mu=0, \sigma)
			\end{array}\right. \\
			R_{\text {collision }}(t)=\left\{\begin{array}{ll}
				-1, \text { if } d_{\text {min }}<d_{\text {collision }} \\
				0, \text { else }
			\end{array}\right\} \\
			R_{\text {total }}=w_1 * R_{\text {goal }}(t)+w_2 * R_{\text {smooth }}(t)+w_3 \\
			* R_{\text {collision }}(t)
		\end{gathered}
		\label{eq:reward}
	\end{small}
\end{equation}
Where $t$ denotes the step number, $\mathcal{N}$ denotes the gaussian function, $p_t$,$p_{goal}$ denotes the current position, and the position of the goal after the agent takes an action, respectively. $\boldsymbol {a_{long}}$ and $\boldsymbol {\delta}$ denotes the longitudinal acceleration and front wheel angle increment of this step, respectively, $d_{min}$ denotes the minimum distance of collision, $w_{i,i=1,2,3}$ is the weight applied to adjust the bias against or toward compatibilities of multi-agent’s predicted trajectories, we set them to $w_1=0.6$,$w_2=0.1$,$w_3=0.5$ empirically. 

It’s worth noting that from the perspective of motion planning, the optimal solution is evidently nonunique, therefore a ground truth trajectory is merely a selection from multiple feasible and optimal modalities. Given that, we calculate the rewards with Gaussian distribution, to avoid overfitting. 
\begin{figure*}
	\centering
	\includegraphics[width=0.9\linewidth]{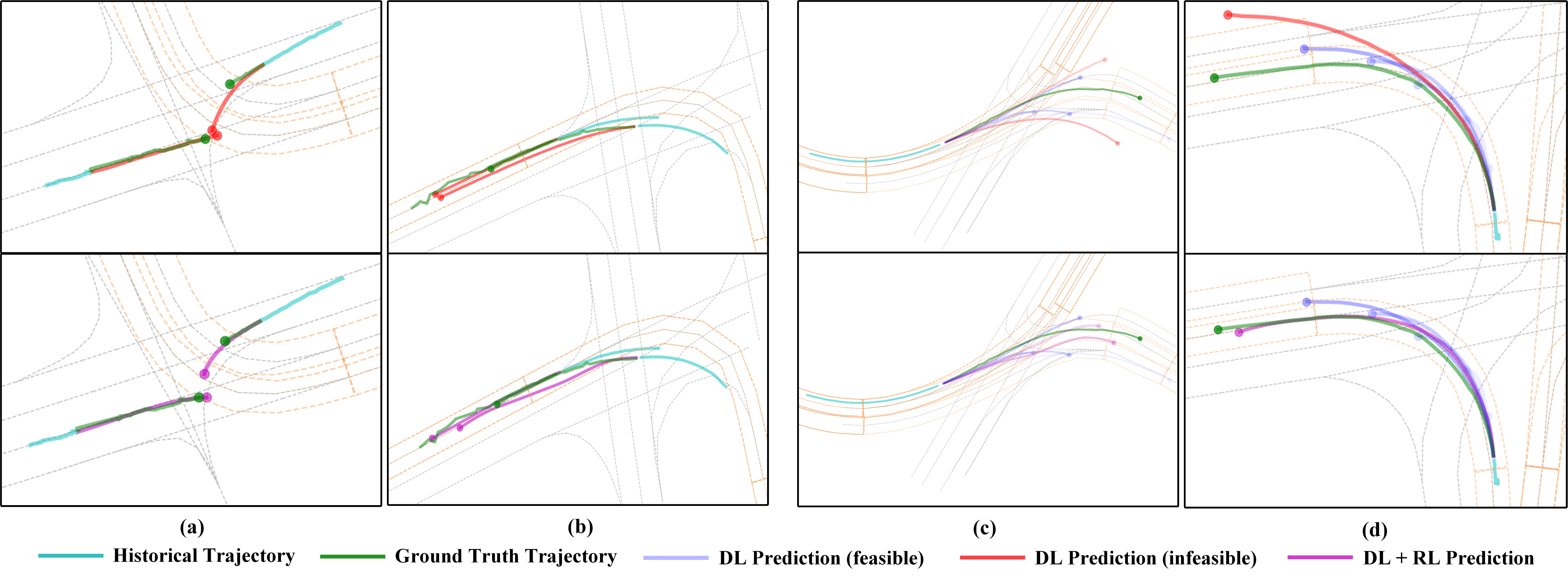}
	\caption{Visualized predictions that demonstrates the effects of the RL stage.}
	\label{fig:Fig4}
\end{figure*}
\section{Experiments}
\subsection{Dataset and Metric}
We evaluate our method on the well-established public dataset: Argoverse Forecasting Dataset \cite{Argoverse}. 

\textbf{Argoverse Motion Forecasting Dataset.} A motion prediction dataset with high-definition (HD) map data and trajectory data sampled at 10 Hz. The dataset contains over 250,000 scenarios. The training, validation, and test sets contain 205942, 39472, and 78143 sequences, respectively. The prediction task is to use the past 2s of history trajectory and the map to predict the next 3s.

\textbf{Metrics.}We evaluate our model in terms of the widely used minimum Average Displacement Error (minADE), minimum Final Displacement Error (minFDE), Miss Rate (MR) and Drivable Area Compliance (DAC).
\subsection{Implement Details}
The heterogeneous graph transformer consists of 1 layer of scene context encoding for local regions with a radius of 50 meters, 3 layers of temporal embedding for agents’ motion states, 3 layers of global interaction encoding and 1 layer of MLP decoder that predicts the key future positions. It is trained 64 epochs on 4 RTX 3090 GPU, with other settings and hyperparameters the same as adopted in HiVT\cite{Zhou2022}. We build a series of sub-scenes as the environment space of the RL model using the ground truth key positions and related lane nodes. During a period of interaction steps, the agents' motion state and their neighboring local context are updated as agents executing every action. The transformer module used in the refined PPO has 4 attention heads, 2 encoder layers and 2 decoder layers with 64 feature dimensions. For training of PPO, we adopt the Adam optimizer with learning rate of 1e-4, the total step size of 512, a buffer size of 8092, the mini-batch size of 1024, the discount factor $\gamma$ = 0.95, the GAE-lamba = 0.97 and the clipped parameter $\varepsilon$ = 0.2.
\subsection{Results}

\textbf{Comparison with State-of-the-art}

The results of comparing our work with SOTA methods on the Argoverse benchmark are presented in \Cref{tab_1}. Our hybrid framework matches the SOTA methods across all metrics and performs best in the term of DAC, which demonstrates the strength of our hybrid framework lies in the ability to produce feasible and plausible predictions, which are scene-wise compatible and compliant to complex constraints, in highly interactive scenes (agent-to-agent and lane-to-agent).

Although a significant overall enhancement on the prediction performance is not observed since highly interactive scenes are sparse in the benchmark, it's of great importance from the perspective of the downstream motion planner of the ego-agent. In Fig. \ref{fig:Fig4}, we visualize the predictions in various scenes where the DL model fails to generate plausible predictions to intuitively demonstrate the effects of our hybrid learning model. The top row of Fig. \ref{fig:Fig4}(a) and Fig. \ref{fig:Fig4}(b) shows cases where the DL stage alone predicts incompatible trajectories between two agents. The top row of Fig. \ref{fig:Fig4}(c) and Fig. \ref{fig:Fig4}(d) show cases where the DL stage alone produces predictions that clearly violate the constraints imposed by the scene context and traffic rules. By comparison, the bottom row shows the feasible predicted trajectories of our proposal in the same scenes. 

It's also worth noting that our RL stage model is trained using only 1\% of the training set, showing great generalization ability.

\textbf{Ablation Study}

\begin{table}
	\begin{center}
		\resizebox{.95\columnwidth}{!}{
			\begin{tabular}{|c|c|c|c|c|}
				\hline
				\begin{tabular}[c]{@{}c@{}}Heterogeous\\Node attr.\end{tabular}  & \begin{tabular}[c]{@{}c@{}}Stacking Aggr.\\+ Comb.\end{tabular}  & minADE & minFDE & MR            \\
				\hline\hline
				/                  &/                          & 0.735          & 1.160           & 0.124 \\
				Lane/Agent         &/                                    & 0.727          & 1.149           & 0.122          \\
				L/R/F/B            &Lane/Agent                  				     & 0.712          & 1.138           & 0.121          \\
				Lane/Agent         &L/R/F/B                                   & \textbf{0.709}          & \textbf{1.114}           & \textbf{0.117}           \\
				\hline
		\end{tabular}}
	\end{center}
	\caption{Prediction results with different encoding methods for heterogeneous context in the DL stage on the Argoverse validation set. The first column describes the type of encoded heterogeneity that is appended to the source node's attributes. The second column describes the heterogeneity upon which multiple Transformers are stacked.}
	\label{tab_2}
\end{table}

{\bf In the DL stage}, we compare different methods of encoding the heterogeneous local context, and present the results in \Cref{tab_2}. The discoveries are two-fold: First, considering heterogeneities is contributive in general. Second, different aspects of heterogeneity, e.g., traffic element types and directional interactions, should be handled differently. Since heterogenous traffic elements such as lanes and agents interact with the target agent simultaneously in a coupling manner, the encoded element type should be appended as a source node attribute for the Transformer to aggregate the coupling influence of heterogeneous context. By comparison, directional heterogeneity could be handled through decomposition followed by a weighted addition, which is also in line with intuition.

{\bf In the RL stage}, we analyze the effects of replacing the NN layers in the vanilla PPO with Transformer, and also the effects of incorporating the vehicle kinematics model into PPO. The results validate the effectiveness of substituting the NN layers with Transformer to model the interaction between agents and the sub-scene context natively with cross-attention. The performance gains of incorporating the vehicle kinematics model into the PPO framework are significant since trajectories compliant to vehicle kinematical constraints are more feasible and better resembles real-life trajectories. It's also discovered that the introduction of the kinematic model speeds up the training process significantly.
\begin{table}
	\begin{center}
		\resizebox{.95\columnwidth}{!}{
			\begin{tabular}{|c|c|c|c|c|}
				\hline
				\begin{tabular}[c]{@{}c@{}}Vanilla-\\PPO\end{tabular}     & \begin{tabular}[c]{@{}c@{}}Transformer-\\PPO\end{tabular}    &\begin{tabular}[c]{@{}c@{}}Kinematic-\\model\end{tabular}  & minADE     & minFDE  \\
				\hline\hline
				\checkmark      &                    &                 & 0.32       & 0.63    \\
				\checkmark      &                    & \checkmark      & 0.23       & 0.50    \\
				& \checkmark         &                 & 0.29       & 0.53    \\
				& \checkmark         & \checkmark      & \textbf{0.19}       & \textbf{0.43}    \\
				\hline
		\end{tabular}}
	\end{center}
	\caption{\centering{The RL stage ablation results on validation set.}}
	\label{tab_3}
\end{table}
\section{Conclusion}
\label{Conclusion}

In this study, we propose a hybrid framework of DL and RL to encode multi-scale interactions for robust multi-modal trajectory prediction. The DL-based stage encodes interactions at the global and intermediate levels with heterogenous Transformers, producing key future positions that represent the agents' multi-modal intentions. The RL-based stage utilizes the predicted key positions as the basis of sub-scene division and transforms the task of predicting trajectories into the procedure of probabilistic motion planning influenced by microscopic interactions with the local scene and subject to vehicle kinematical constraints. Experiments show that our novel hybrid learning framework achieves SOTA performance on the Argoverse forecasting benchmark and produces more feasible and plausible predictions in highly interactive scenes.

\ifCLASSOPTIONcaptionsoff
  \newpage
\fi

\bibliographystyle{IEEEtran}
\bibliography{egbib}

\vfill

\end{document}